\documentclass[runningheads]{llncs}
\usepackage[T1]{fontenc}
\usepackage{graphicx}

\usepackage{amsmath,amssymb}
\usepackage{bm}
\usepackage{xcolor}
\usepackage{colortbl}
\usepackage{booktabs}
\usepackage{multirow}
\usepackage{url}

\def\overview{
\begin{figure}[t]
\centering
\includegraphics[width=0.6\linewidth]{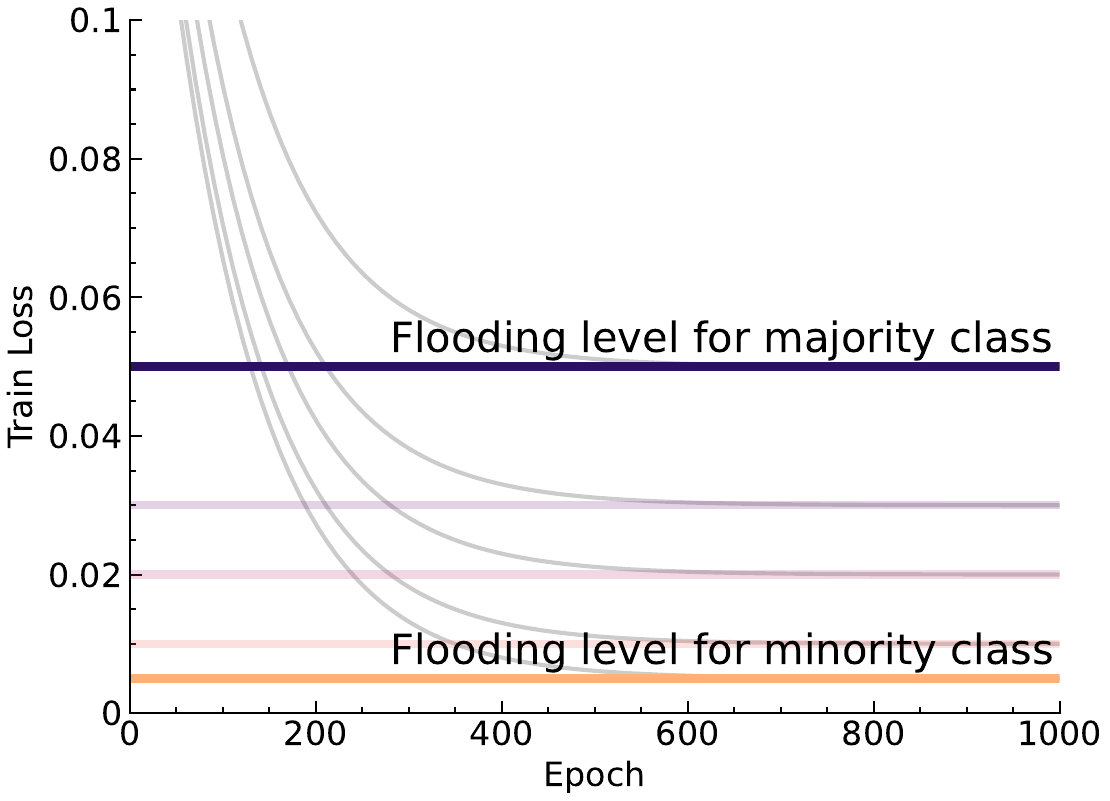}
\caption{Concept of Class-wise Flooding. To improve learning from imbalanced data, we propose a Flooding regularization scheme that assigns a separate Flooding Level to each class. These class-wise Flooding Levels are determined based on the relative frequency of samples in each class.}
\label{fig:overview}
\end{figure}
}

\newcommand{\colwidthtraintestcurve}{0.31\textwidth}
\def\figtraintestcurvelow{
\begin{figure*}[t]
\centering
\begin{minipage}[b]{\colwidthtraintestcurve}{
\includegraphics[width=1.0\linewidth]{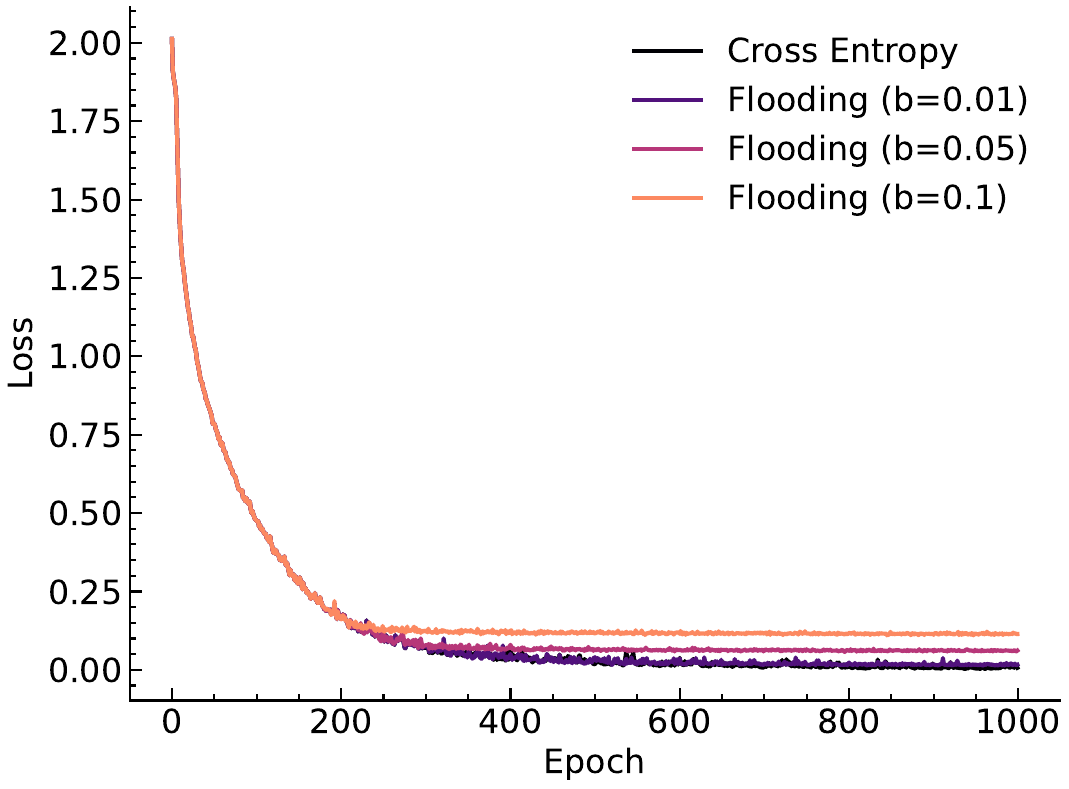}
}
\end{minipage}
\hfill
\begin{minipage}[b]{\colwidthtraintestcurve}{
\includegraphics[width=1.0\linewidth]{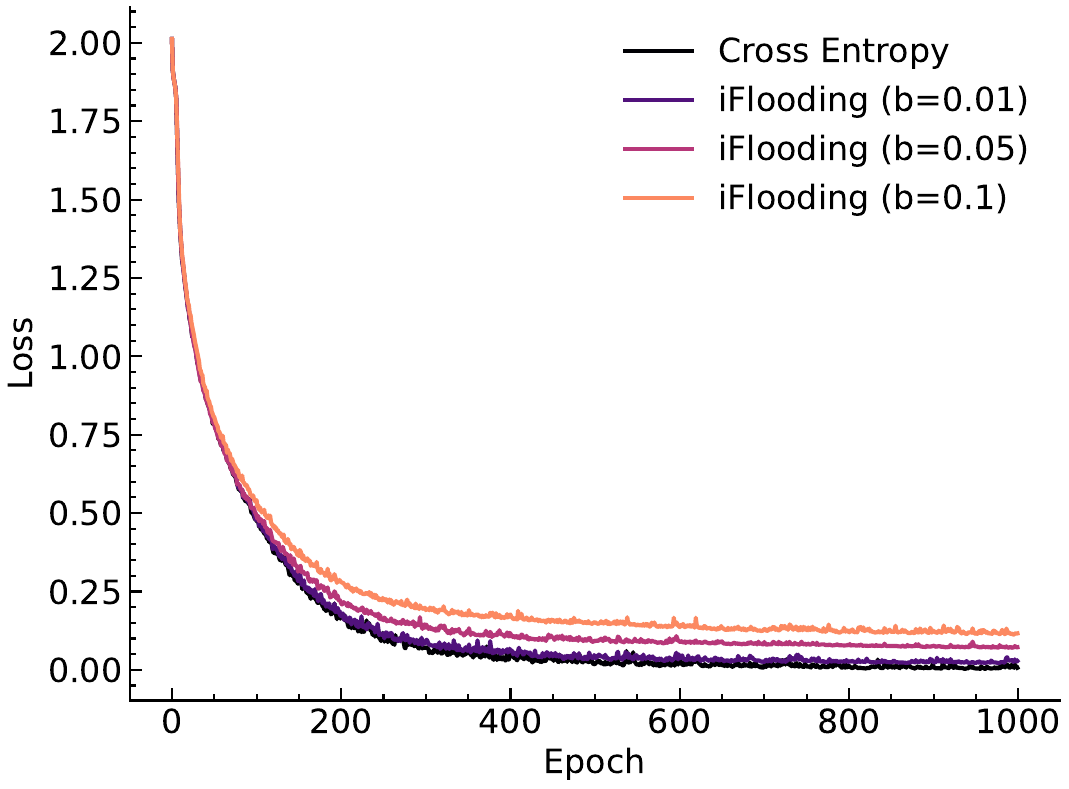}
}
\end{minipage}
\hfill
\begin{minipage}[b]{\colwidthtraintestcurve}{
\includegraphics[width=1.0\linewidth]{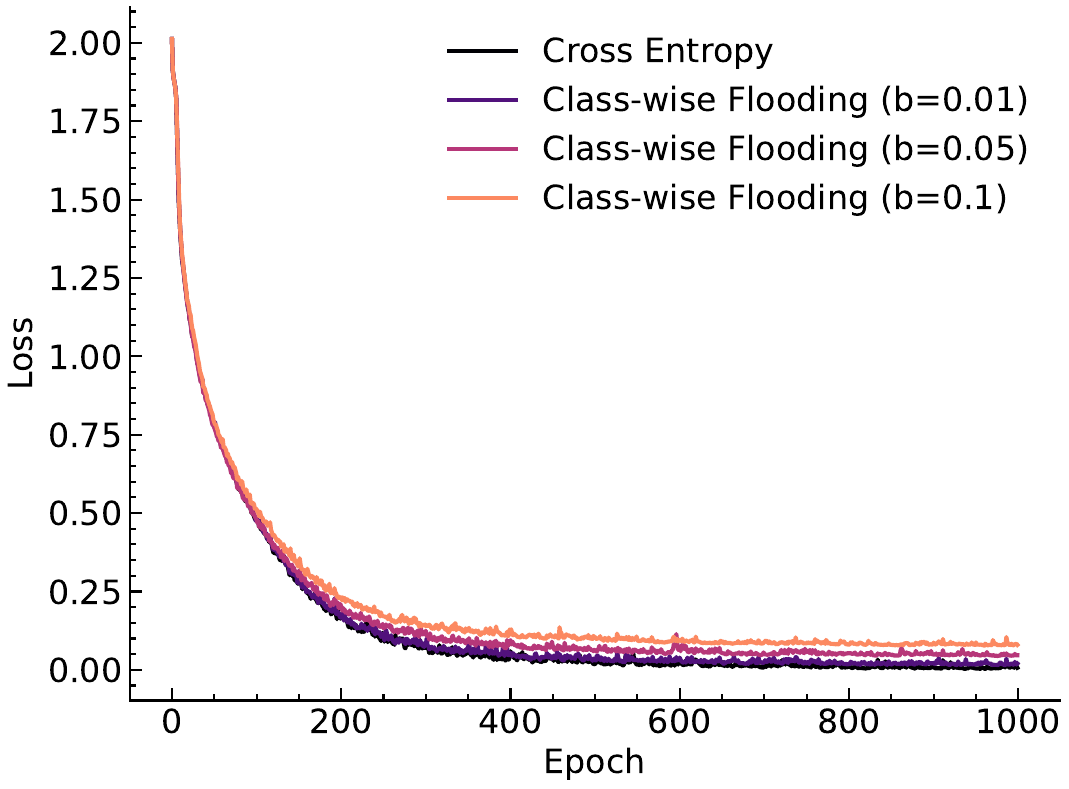}
}
\end{minipage}
\vspace{0.5cm}
\begin{minipage}[b]{\colwidthtraintestcurve}{
\includegraphics[width=1.0\linewidth]{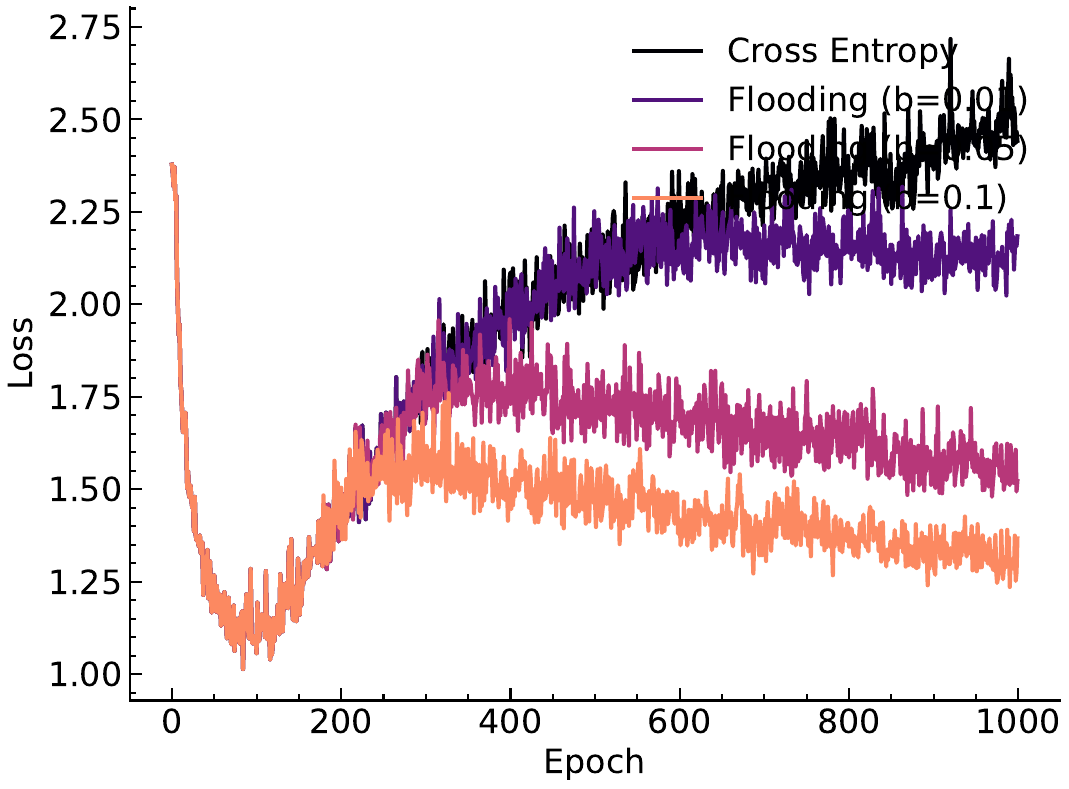}
\makebox[\linewidth]{\footnotesize Flooding}
}
\end{minipage}
\hfill
\begin{minipage}[b]{\colwidthtraintestcurve}{
\includegraphics[width=1.0\linewidth]{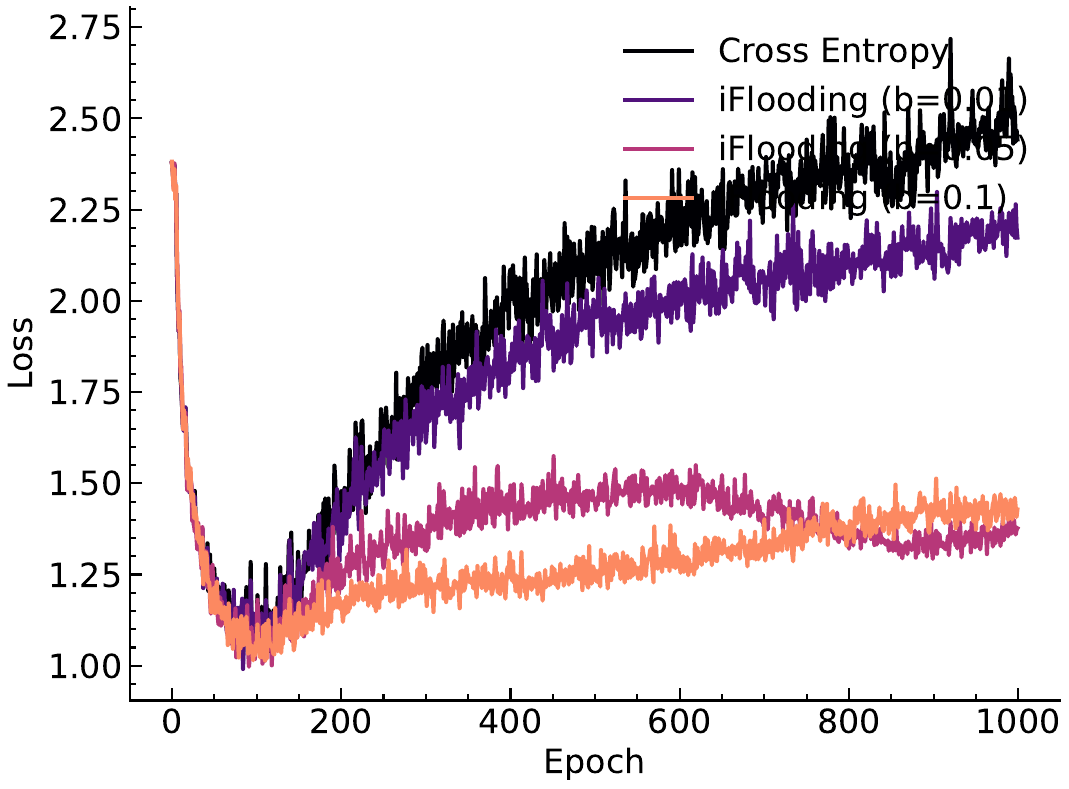}
\makebox[\linewidth]{\footnotesize iFlooding}
}
\end{minipage}
\hfill
\begin{minipage}[b]{\colwidthtraintestcurve}{
\includegraphics[width=1.0\linewidth]{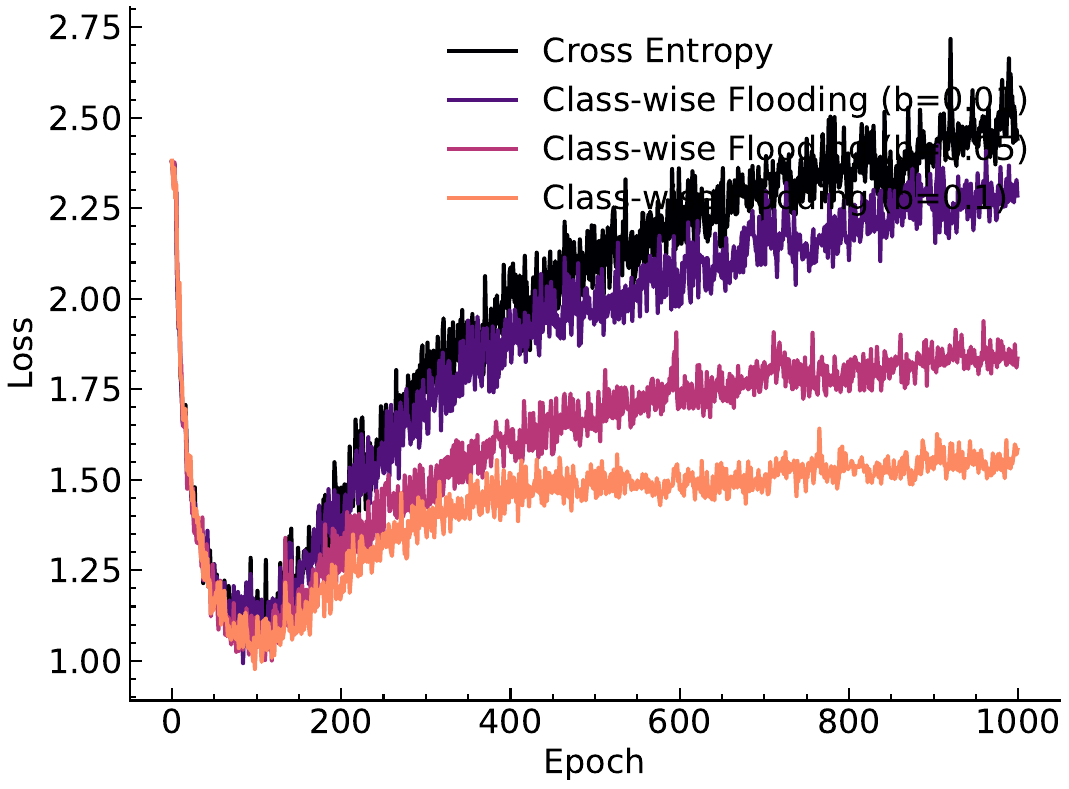}
\makebox[\linewidth]{\footnotesize Class-wise Flooding}
}
\end{minipage}
\caption{Visualization of training and validation loss for Flooding, iFlooding, and Class-wise Flooding regularization on \texttt{CIFAR10-LT} with imbalance ratio $\rho = 10$.}
\label{fig:traintestcurve10}
\end{figure*}
}

\def\figclasswisetraintestcurvelow{
\begin{figure*}[t!]
\centering
\begin{minipage}[b]{\colwidthtraintestcurve}{
\includegraphics[width=1.0\linewidth]{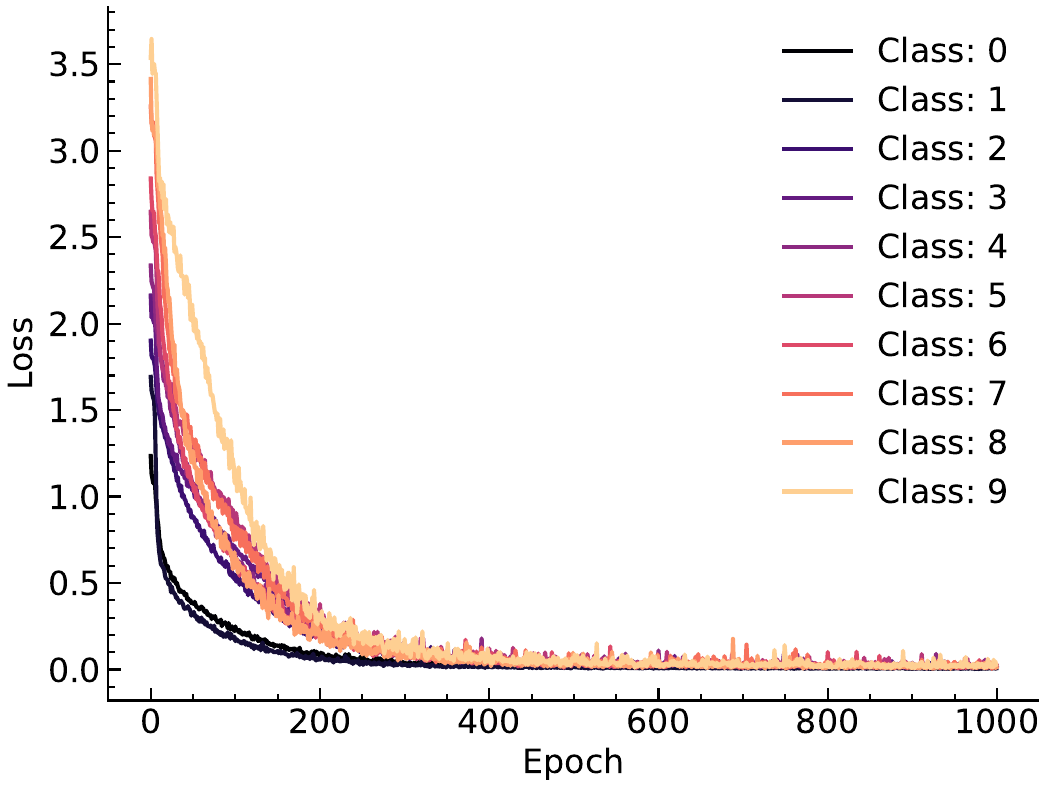}
}
\end{minipage}
\hfill
\begin{minipage}[b]{\colwidthtraintestcurve}{
\includegraphics[width=1.0\linewidth]{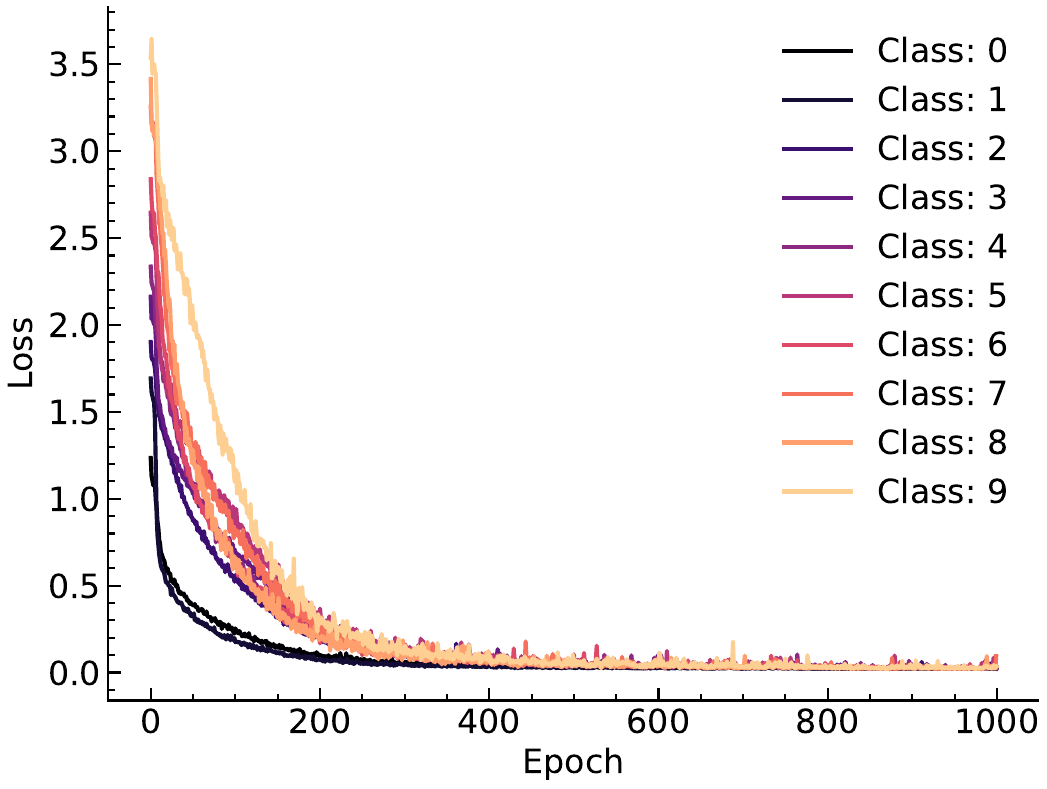}
}
\end{minipage}
\hfill
\begin{minipage}[b]{\colwidthtraintestcurve}{
\includegraphics[width=1.0\linewidth]{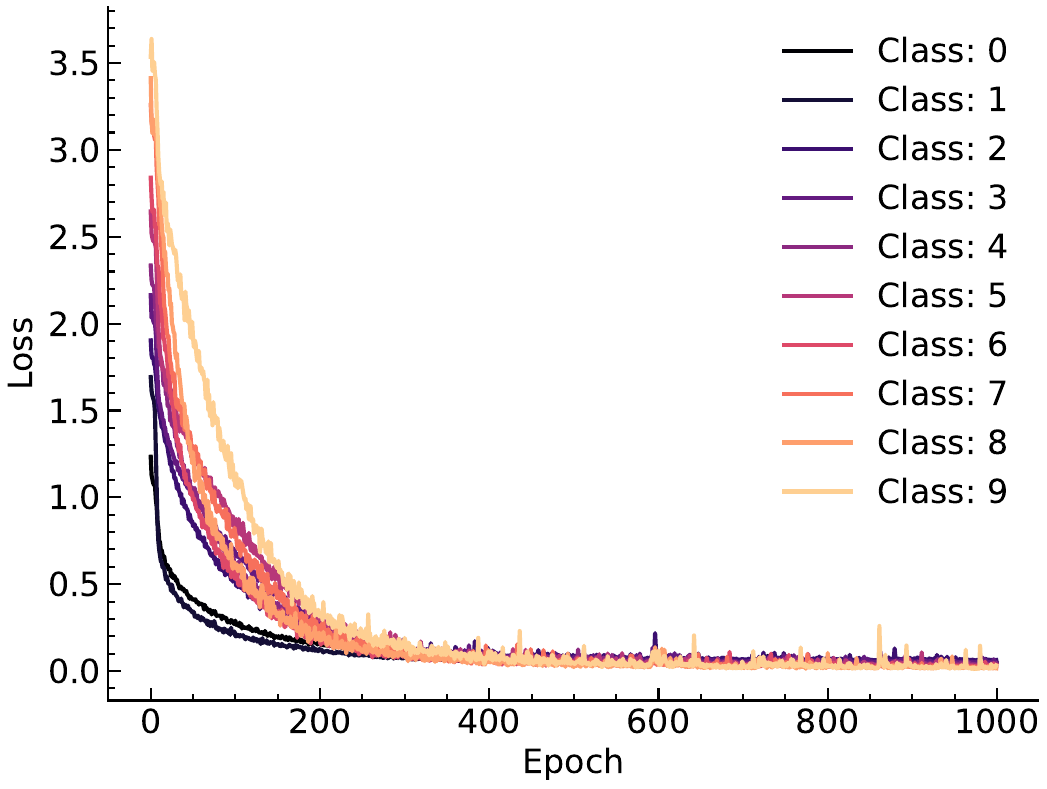}
}
\end{minipage}
\vspace{0.5cm}
\begin{minipage}[b]{\colwidthtraintestcurve}{
\includegraphics[width=1.0\linewidth]{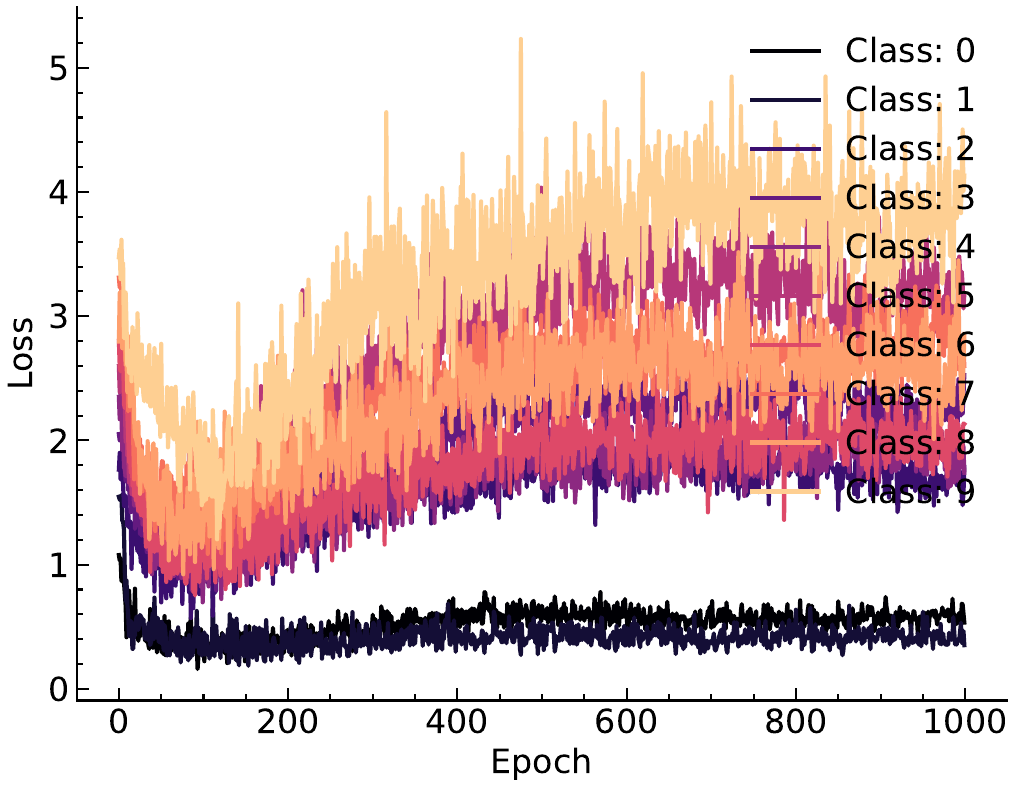}
\makebox[\linewidth]{\footnotesize Flooding}
}
\end{minipage}
\hfill
\begin{minipage}[b]{\colwidthtraintestcurve}{
\includegraphics[width=1.0\linewidth]{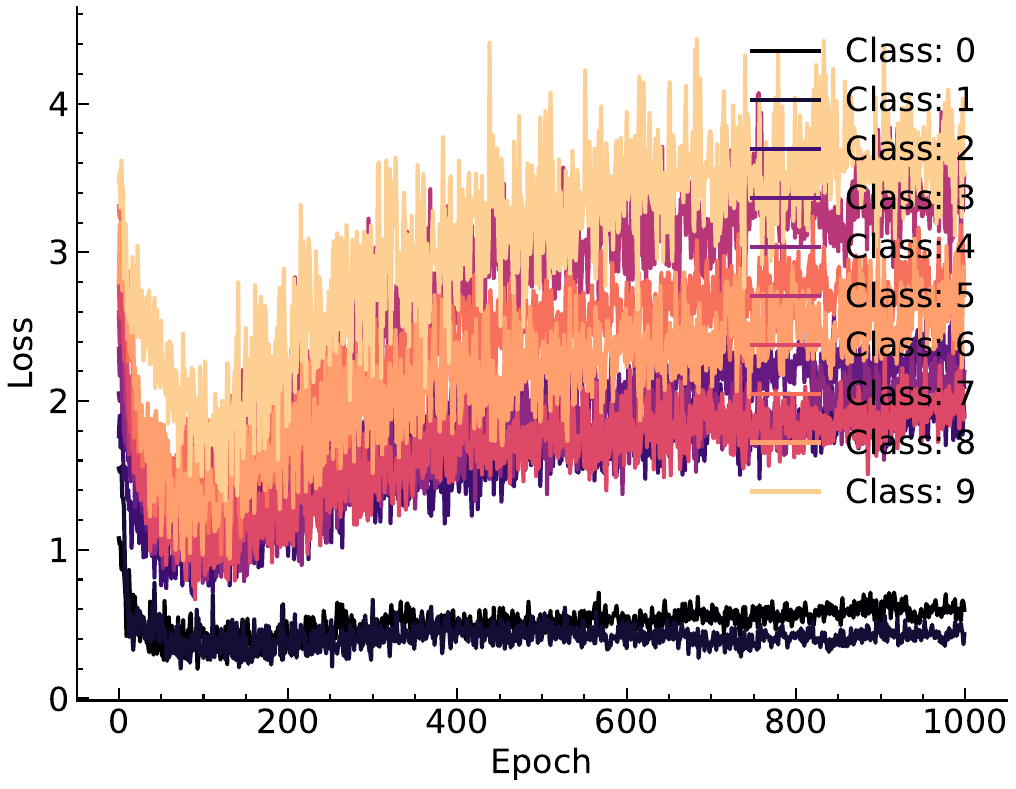}
\makebox[\linewidth]{\footnotesize iFlooding}
}
\end{minipage}
\hfill
\begin{minipage}[b]{\colwidthtraintestcurve}{
\includegraphics[width=1.0\linewidth]{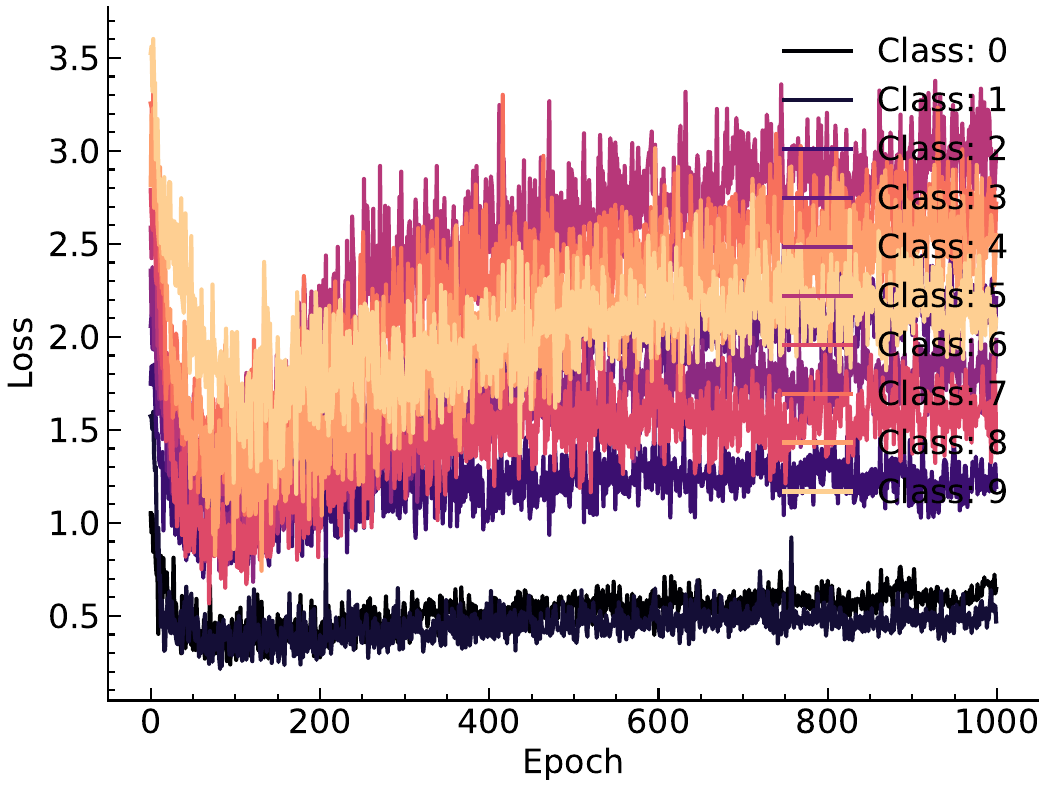}
\makebox[\linewidth]{\footnotesize Class-wise Flooding}
}
\end{minipage}
\caption{Class-wise visualization of training and validation loss for Flooding, iFlooding, and Class-wise Flooding regularization on \texttt{CIFAR10-LT} with imbalance ratio $\rho = 10$.}
\label{fig:classwisetraintestcurve10}
\end{figure*}
}

\def\figclasswisetraintestcurvehigh{
\begin{figure*}[t!]
\centering
\begin{minipage}[b]{\colwidthtraintestcurve}{
\includegraphics[width=1.0\linewidth]{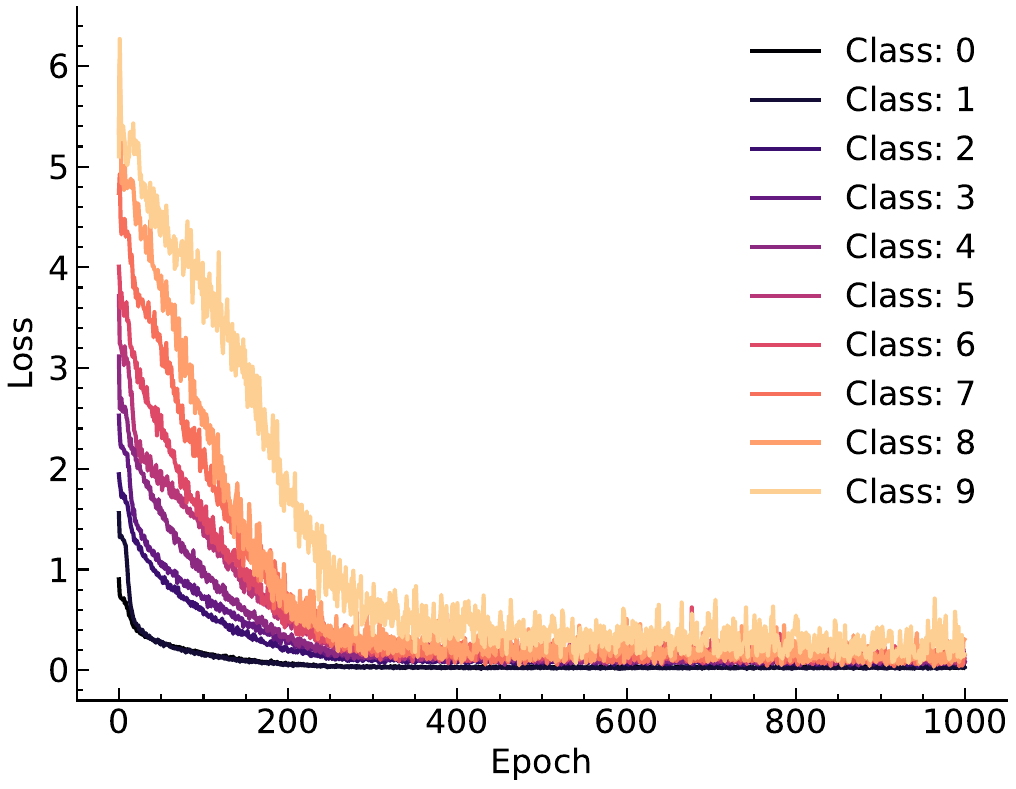}
}
\end{minipage}
\hfill
\begin{minipage}[b]{\colwidthtraintestcurve}{
\includegraphics[width=1.0\linewidth]{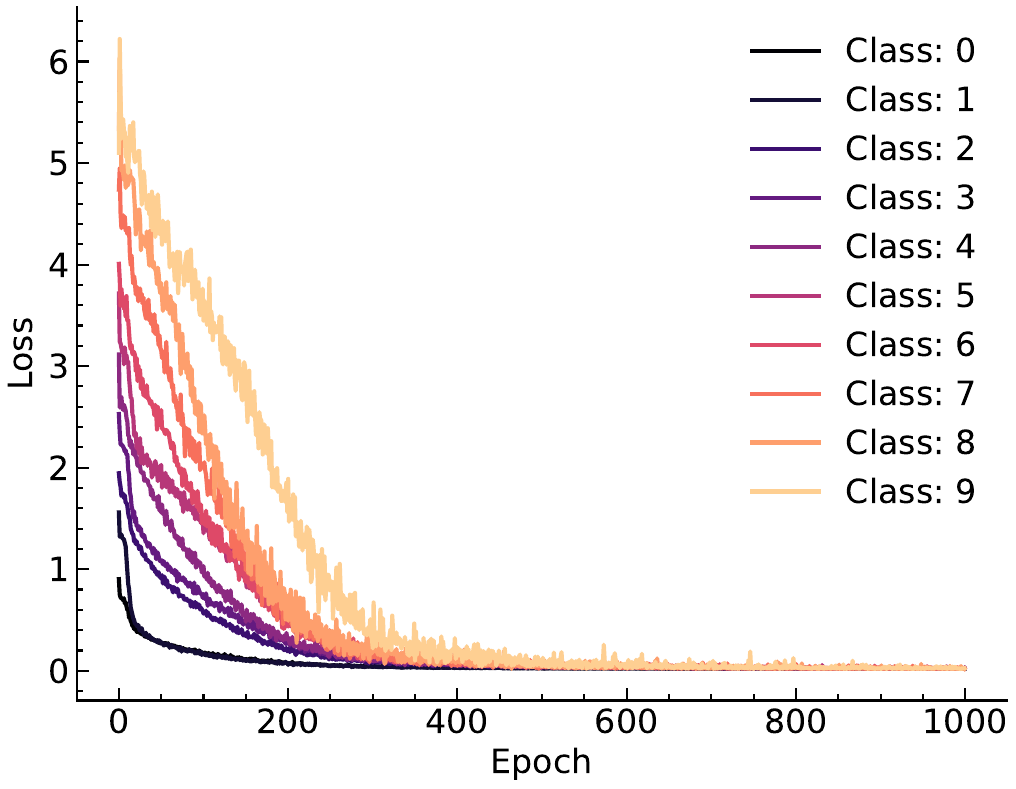}
}
\end{minipage}
\hfill
\begin{minipage}[b]{\colwidthtraintestcurve}{
\includegraphics[width=1.0\linewidth]{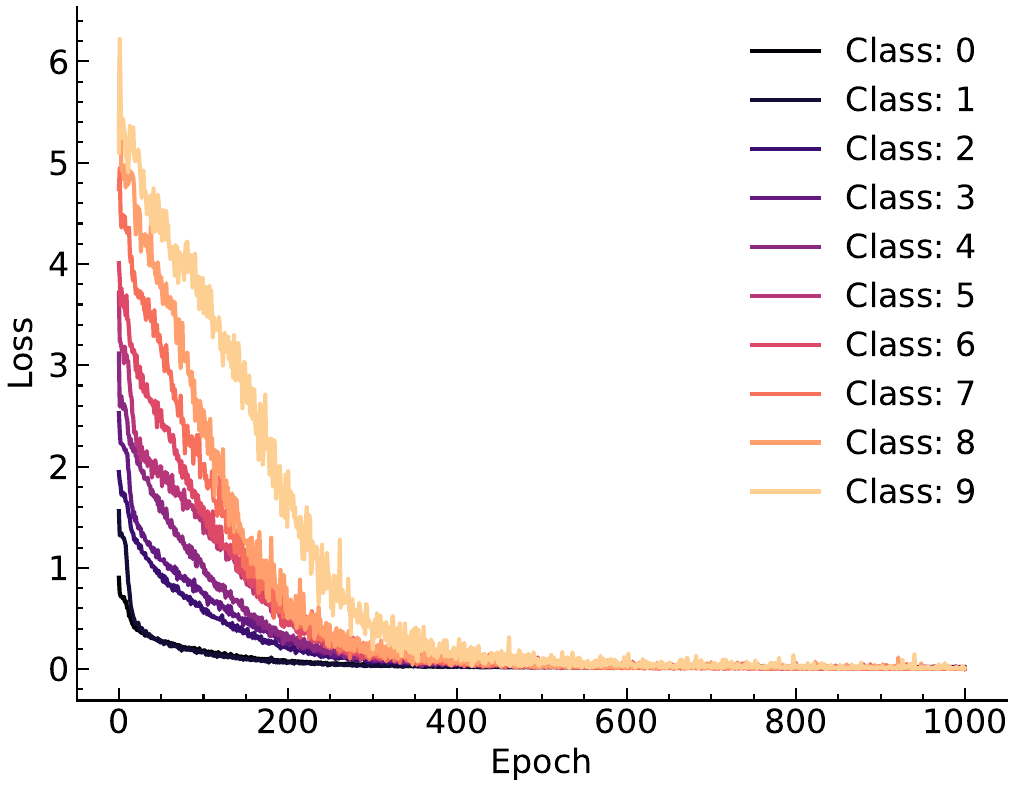}
}
\end{minipage}
\vspace{0.5cm}
\begin{minipage}[b]{\colwidthtraintestcurve}{
\includegraphics[width=1.0\linewidth]{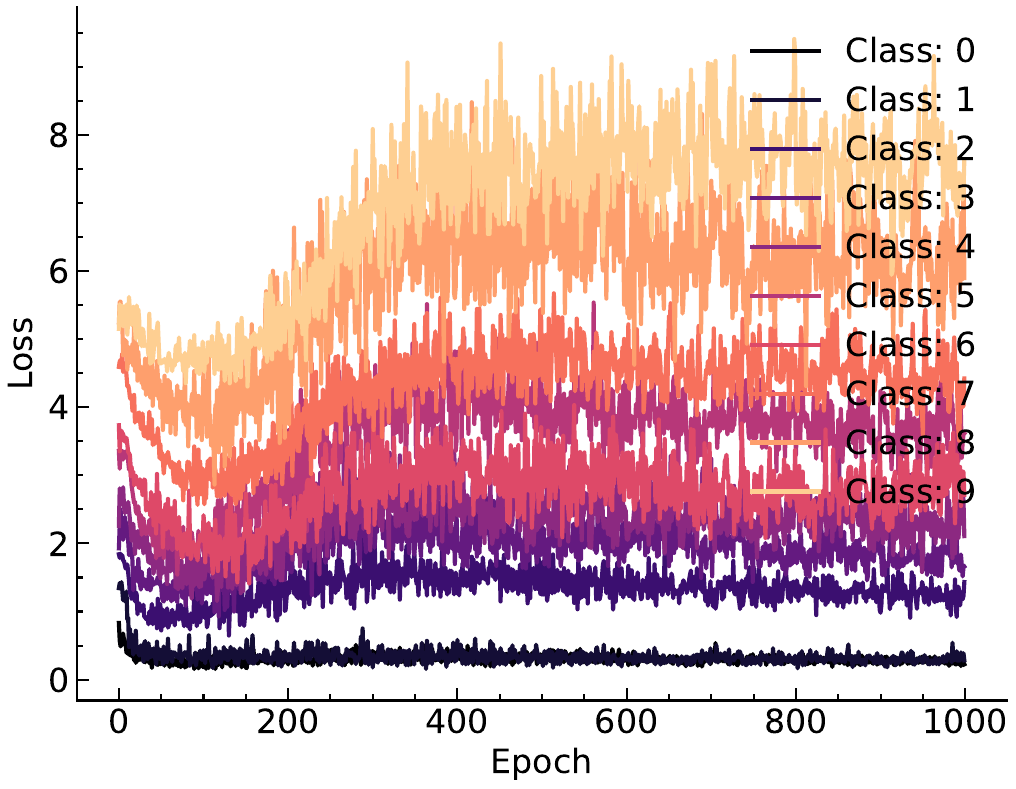}
\makebox[\linewidth]{\footnotesize Flooding}
}
\end{minipage}
\hfill
\begin{minipage}[b]{\colwidthtraintestcurve}{
\includegraphics[width=1.0\linewidth]{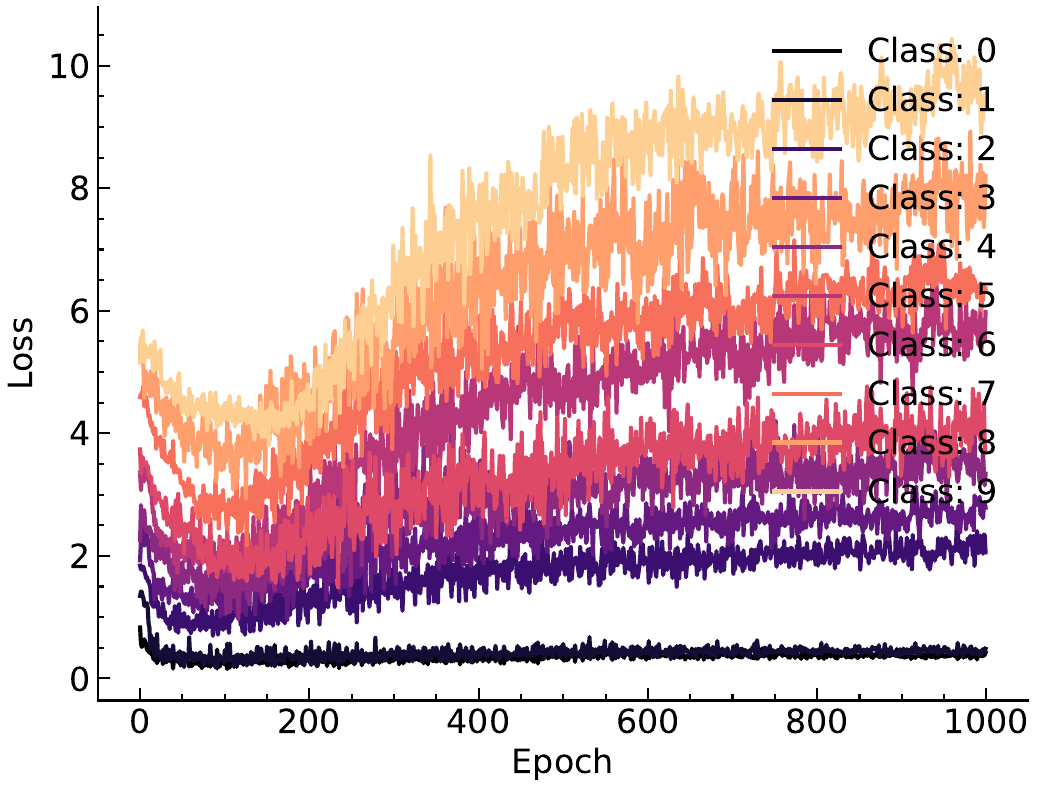}
\makebox[\linewidth]{\footnotesize iFlooding}
}
\end{minipage}
\hfill
\begin{minipage}[b]{\colwidthtraintestcurve}{
\includegraphics[width=1.0\linewidth]{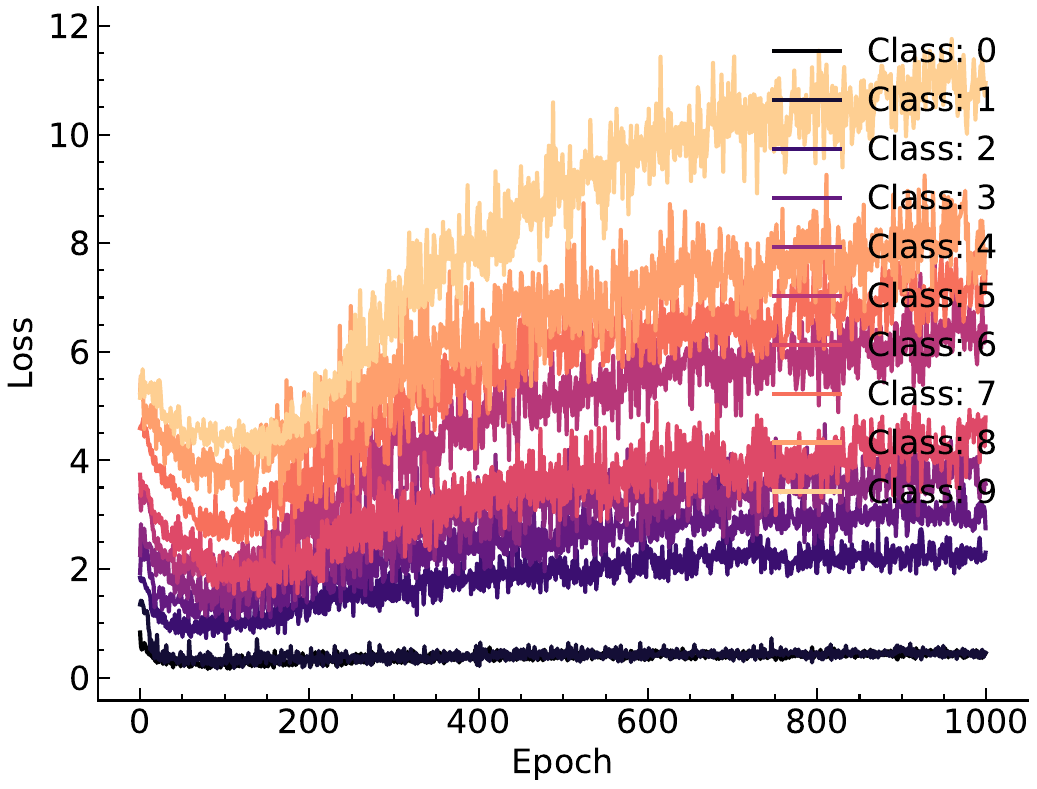}
\makebox[\linewidth]{\footnotesize Class-wise Flooding}
}
\end{minipage}
\caption{Class-wise visualization of training and validation loss for Flooding, iFlooding, and Class-wise Flooding regularization on \texttt{CIFAR10-LT} with imbalance ratio $\rho = 100$.}
\label{fig:classwisetraintestcurve100}
\end{figure*}
}
\def\lowimbalancecifarten{
\begin{table*}[t]
\centering
\caption{Class-wise and average accuracy on \texttt{CIFAR10-LT} with imbalance ratio $\rho = 10$.}
\resizebox{\linewidth}{!}{
\begin{tabular}{lccccccccccc}
\toprule[0.4mm]
\multicolumn{1}{c}{} & \multicolumn{10}{c}{\textbf{Class}} & \multicolumn{1}{c}{} \\
\cmidrule(lr){2-11}
\textbf{Method} & 0 & 1 & 2 & 3 & 4 & 5 & 6 & 7 & 8 & 9 & \textbf{Avg} \\
\midrule
Cross Entropy & $89.5_{\pm 1.53}$ & $91.2_{\pm 1.27}$ & $71.23_{\pm 2.9}$ & $63.53_{\pm 1.68}$ & $71.97_{\pm 1.93}$ & $57.6_{\pm 1.44}$ & $74.67_{\pm 1.64}$ & $\mathbf{66.6_{\pm 1.31}}$ & $69.7_{\pm 1.68}$ & $53.47_{\pm 4.08}$ & $70.95_{\pm 0.63}$ \\
\qquad w/ Class Weights & $87.27_{\pm 0.19}$ & $89.37_{\pm 1.41}$ & $69.9_{\pm 1.35}$ & $63.4_{\pm 1.87}$ & $67.5_{\pm 0.83}$ & $57.2_{\pm 1.51}$ & $72.0_{\pm 1.13}$ & $63.2_{\pm 1.88}$ & $\mathbf{70.73_{\pm 0.45}}$ & $\mathbf{59.0_{\pm 2.63}}$ & $69.96_{\pm 0.79}$ \\
\qquad w/ Focal Weights & $84.53_{\pm 0.87}$ & $86.57_{\pm 0.26}$ & $66.63_{\pm 1.4}$ & $60.4_{\pm 1.1}$ & $64.97_{\pm 1.48}$ & $56.17_{\pm 2.9}$ & $66.17_{\pm 2.08}$ & $61.33_{\pm 1.63}$ & $60.93_{\pm 3.71}$ & $44.5_{\pm 1.2}$ & $65.22_{\pm 0.65}$ \\
\qquad w/ Flood & $88.17_{\pm 1.23}$ & $91.47_{\pm 0.46}$ & $73.23_{\pm 1.16}$ & $62.17_{\pm 0.7}$ & $\mathbf{72.93_{\pm 2.63}}$ & $57.1_{\pm 2.3}$ & $75.03_{\pm 2.95}$ & $65.57_{\pm 1.37}$ & $70.17_{\pm 1.3}$ & $55.07_{\pm 2.38}$ & $71.09_{\pm 0.49}$ \\
\qquad w/ iFlood & $89.1_{\pm 1.42}$ & $91.17_{\pm 0.87}$ & $73.43_{\pm 1.86}$ & $\mathbf{64.53_{\pm 1.4}}$ & $71.13_{\pm 1.87}$ & $56.03_{\pm 2.75}$ & $73.13_{\pm 3.01}$ & $64.73_{\pm 0.83}$ & $69.6_{\pm 1.59}$ & $55.07_{\pm 1.86}$ & $70.79_{\pm 0.38}$ \\
\rowcolor{black!7}\qquad w/ Class-wise Flood & $\mathbf{90.23_{\pm 0.66}}$ & $\mathbf{91.9_{\pm 0.65}}$ & $\mathbf{74.4_{\pm 0.93}}$ & $64.17_{\pm 2.26}$ & $69.7_{\pm 0.85}$ & $\mathbf{58.63_{\pm 0.46}}$ & $\mathbf{75.47_{\pm 1.26}}$ & $66.1_{\pm 0.92}$ & $69.9_{\pm 1.81}$ & $54.73_{\pm 1.4}$ & $\mathbf{71.52_{\pm 0.34}}$ \\
\bottomrule[0.4mm]
\end{tabular}
\label{tab:cifar10rho10}
}
\end{table*}
}

\def\highimbalancecifarten{
\begin{table*}[t]
\centering
\caption{Class-wise and average accuracy on \texttt{CIFAR10-LT} with imbalance ratio $\rho = 100$.}
\resizebox{\linewidth}{!}{
\begin{tabular}{lccccccccccc}
\toprule[0.4mm]
\multicolumn{1}{c}{} & \multicolumn{10}{c}{\textbf{Class}} & \multicolumn{1}{c}{} \\
\cmidrule(lr){2-11}
\textbf{Method} & 0 & 1 & 2 & 3 & 4 & 5 & 6 & 7 & 8 & 9 & \textbf{Avg} \\
\midrule
Cross Entropy & $92.7_{\pm 0.92}$ & $92.0_{\pm 0.92}$ & $68.9_{\pm 2.59}$ & $56.0_{\pm 3.36}$ & $55.97_{\pm 2.45}$ & $41.1_{\pm 4.92}$ & $56.23_{\pm 2.58}$ & $36.83_{\pm 4.12}$ & $22.07_{\pm 5.36}$ & $7.63_{\pm 2.52}$ & $52.94_{\pm 1.16}$ \\
\qquad w/ Class Weights & $75.87_{\pm 2.33}$ & $79.07_{\pm 1.28}$ & $44.47_{\pm 2.5}$ & $36.2_{\pm 1.15}$ & $50.47_{\pm 2.4}$ & $\mathbf{47.73_{\pm 0.69}}$ & $\mathbf{60.17_{\pm 5.75}}$ & $\mathbf{45.37_{\pm 1.46}}$ & $\mathbf{33.43_{\pm 6.58}}$ & $\mathbf{19.23_{\pm 1.13}}$ & $49.2_{\pm 0.65}$ \\
\qquad w/ Focal Weights & $88.5_{\pm 1.27}$ & $89.93_{\pm 2.38}$ & $63.73_{\pm 1.64}$ & $58.43_{\pm 0.42}$ & $47.97_{\pm 6.73}$ & $34.07_{\pm 3.42}$ & $40.87_{\pm 2.31}$ & $33.0_{\pm 3.16}$ & $10.67_{\pm 2.75}$ & $1.3_{\pm 0.71}$ & $46.85_{\pm 0.98}$ \\
\qquad w/ Flood & $91.13_{\pm 0.95}$ & $91.17_{\pm 2.05}$ & $\mathbf{69.8_{\pm 1.31}}$ & $\mathbf{59.17_{\pm 1.19}}$ & $57.83_{\pm 2.19}$ & $38.87_{\pm 3.16}$ & $53.8_{\pm 2.3}$ & $36.5_{\pm 3.75}$ & $23.57_{\pm 6.27}$ & $7.3_{\pm 2.43}$ & $52.91_{\pm 0.79}$ \\
\qquad w/ iFlood & $\mathbf{93.47_{\pm 1.04}}$ & $\mathbf{92.8_{\pm 1.16}}$ & $67.4_{\pm 2.68}$ & $59.07_{\pm 1.39}$ & $58.3_{\pm 2.59}$ & $43.23_{\pm 2.53}$ & $51.43_{\pm 1.17}$ & $36.9_{\pm 1.93}$ & $17.07_{\pm 2.29}$ & $5.97_{\pm 1.55}$ & $52.56_{\pm 0.32}$ \\
\rowcolor{black!7}\qquad w/ Class-wise Flood & $90.93_{\pm 1.99}$ & $91.87_{\pm 1.45}$ & $68.6_{\pm 0.7}$ & $57.3_{\pm 0.37}$ & $\mathbf{61.13_{\pm 2.74}}$ & $40.97_{\pm 4.83}$ & $52.13_{\pm 2.78}$ & $38.97_{\pm 4.26}$ & $20.97_{\pm 2.1}$ & $9.47_{\pm 1.64}$ & $\mathbf{53.23_{\pm 0.75}}$ \\
\bottomrule[0.4mm]
\end{tabular}
\label{tab:cifar10rho100}
}
\end{table*}
}

\def\imbalancecifarhun{
\begin{table}[t]
\centering
\caption{Average accuracy on \texttt{CIFAR100-LT}.}
\begin{tabular}{lccccccccccc}
\toprule[0.4mm]
\multicolumn{1}{c}{} & \multicolumn{2}{c}{$\mathbf{\rho}$} \\
\cmidrule(lr){2-3}
\textbf{Method} & 10 & 100 \\
\midrule
Cross Entropy & $\mathbf{36.9_{\pm 0.52}}$ & $25.5_{\pm 0.14}$ \\
\qquad w/ Class Weights & $33.55_{\pm 0.52}$ & $18.75_{\pm 0.56}$ \\
\qquad w/ Focal Weights & $9.05_{\pm 0.72}$ & $5.03_{\pm 0.2}$ \\
\qquad w/ Flood & $36.16_{\pm 0.72}$ & $25.73_{\pm 0.19}$ \\
\qquad w/ iFlood &  $36.36_{\pm 0.74}$ & $25.66_{\pm 0.21}$ \\
\rowcolor{black!7}\qquad w/ Class-wise Flood & $36.86_{\pm 0.58}$ & $\mathbf{26.39_{\pm 0.41}}$ \\
\bottomrule[0.4mm]
\end{tabular}
\label{tab:cifar100}
\end{table}
}

\def\bbase{
\begin{table}[t]
\centering
\caption{Classification accuracy on \texttt{CIFAR10-LT} with different values of $b_{\mathrm{base}}$.}
\begin{tabular}{lccccccccccc}
\toprule[0.4mm]
\multicolumn{1}{c}{} & \multicolumn{2}{c}{$\mathbf{\rho}$} \\
\cmidrule(lr){2-3}
\textbf{Method} & 10 & 100 \\
\midrule
iFlood ($b=0.01$) & $70.79_{\pm 0.38}$ & $52.56_{\pm 0.32}$ \\
iFlood ($b=0.05$) & $70.16_{\pm 0.53}$ & $51.5_{\pm 0.29}$ \\
iFlood ($b=0.1$) & $70.44_{\pm 0.45}$ & $51.0_{\pm 0.11}$ \\
\rowcolor{black!7}Class-wise Flood ($b_\mathrm{base}=0.01$) & $70.86_{\pm 0.87}$ & $\mathbf{53.23_{\pm 0.75}}$ \\
\rowcolor{black!7}Class-wise Flood ($b_\mathrm{base}=0.05$) & $\mathbf{71.52_{\pm 0.34}}$ & $52.18_{\pm 0.3}$ \\
\rowcolor{black!7}Class-wise Flood ($b_\mathrm{base}=0.1$) & $71.03_{\pm 0.39}$ & $52.27_{\pm 0.78}$ \\
\bottomrule[0.4mm]
\end{tabular}
\label{tab:bbase}
\end{table}
}

\makeatletter
\newcommand{\samethanks}[1][\value{footnote}]{\footnotemark[#1]}
\makeatother

\begin{document}
\title{Class-wise Flooding Regularization for Imbalanced Image Classification}
\titlerunning{Class-wise Flooding Regularization}

\author{Hiroaki Aizawa\thanks{These authors contributed equally to this work.} \and Yuta Naito\samethanks \and Kohei Fukuda}
\authorrunning{H. Aizawa et al.}

\institute{Graduate School of Advanced Science and Engineering\\
Hiroshima University\\
Higashi-Hiroshima, Hiroshima, Japan\\
\email{hiroaki-aizawa@hiroshima-u.ac.jp}}
\maketitle              
\begin{abstract}
The purpose of training neural networks is to achieve high generalization performance on unseen inputs. However, when trained on imbalanced datasets, a model's prediction tends to favor majority classes over minority classes, leading to significant degradation in the recognition performance of minority classes. To address this issue, we propose class-wise flooding regularization, an extension of flooding regularization applied at the class level. Flooding is a regularization technique that mitigates overfitting by preventing the training loss from falling below a predefined threshold, known as the flooding level, thereby discouraging memorization. Our proposed method assigns a class-specific flooding level based on class frequencies. By doing so, it suppresses overfitting in majority classes while allowing sufficient learning for minority classes. We validate our approach on imbalanced image classification. Compared to conventional flooding regularizations, our method improves the classification performance of minority classes and achieves better overall generalization.
\end{abstract}

\section{Introduction}
The goal of training neural networks is to achieve strong generalization to unseen data or domain. To this end, various approaches have been proposed, including regularization techniques such as weight decay~\cite{weight_decay}, dropout~\cite{dropout} and early stopping~\cite{early_stopping}, model architectures such as residual connection~\cite{resnet} and the self-attention mechanism~\cite{transformer}, and flatness-aware optimization methods~\cite{sam}.

In general, generalization performance is influenced not only by the techniques mentioned above but also by the properties of the training data. In particular, when the dataset is imbalanced with uneven class distributions, or contains label noise, the model may suffer from severe performance degradation or unstable training dynamics. Among these, class imbalance is especially important because it frequently arises in practical applications such as medical image diagnosis and fraud detection. When training with imbalanced data, learning tends to be biased toward majority classes, resulting in insufficient learning for minority classes. This bias arises during the sampling of mini-batches, where majority class samples are disproportionately selected. As a result, majority classes tend to converge quickly, whereas minority classes are not sufficiently trained. This biased training process causes the model to become skewed in its predictions toward the majority, which makes it difficult to accurately recognize minority classes.

\overview

To address this issue, we investigate a regularization approach that equalizes training progress across classes by controlling the loss dynamics during optimization. In particular, we focus on Flooding regularization~\cite{flood}, which prevents the training loss from decreasing below a predefined threshold, known as the flooding level. By maintaining the training loss at a moderate level, the model mitigates overfitting and memorization to the training data and achieves more robust generalization on unseen samples. Prior work~\cite{soft_ad} has shown, through analysis of update steps and loss landscapes, that Flooding regularization achieves effects comparable to those of Sharpness-Aware Minimization~\cite{sam} and gradient norm regularization~\cite{grad_norm}. Moreover, follow-up studies have proposed instance-wise extensions such as iFlooding~\cite{iflood} and adaptive schemes such as AdaFlood~\cite{adaflood}, which automatically adjust the flooding level using auxiliary data, offering new directions for controlling training dynamics.

Building on these insights, as illustrated in Figure~\ref{fig:overview}, we propose a class-wise Flooding regularization method where the flooding level is weighted according to the sample frequency of each class. In this formulation, higher flooding levels are assigned to majority classes to suppress overfitting caused by rapid convergence, while lower flooding levels are assigned to minority classes to encourage sufficient learning. Through experiments on imbalanced datasets, we compare the proposed method with conventional Flooding and iFlooding regularizations. In addition, we provide quantitative and qualitative analysis of the relationship between class-wise flooding levels, learning curves, per-class classification accuracy under varying imbalance ratios. Experimental results demonstrate that our method achieves superior performance, particularly for minority classes, and contributes to improved overall generalization.

\section{Preliminaries and Related Work}

We consider a $K$-class classification problem, where a neural network is trained using $d$-dimensional input data $\bm{x} \in \mathbb{R}^d$ and its corresponding $K$-dimensional one-hot label vector $\bm{y}$. The classification model $f: \mathbb{R}^d \rightarrow \mathbb{R}^K$ outputs a logit vector:
\begin{align}
f(\bm{x}) = [f_1(\bm{x}), f_2(\bm{x}), \dots, f_K(\bm{x})]
\end{align}
From the logits, a predicted probability distribution is computed using the softmax function, and the cross-entropy loss $\ell_{\mathrm{CE}}$ between the prediction and the ground-truth label is minimized. Specifically, the cross-entropy loss is defined as:
\begin{align}
\ell_{\mathrm{CE}}(f(\bm{x}), \bm{y}) = -\sum_{i=1}^{K} y_i \log \frac{\exp(f_i(\bm{x}))}{\sum_{j=1}^{K} \exp(f_j(\bm{x}))}
\end{align}
In practice, the model parameters are optimized to minimize the empirical loss, which is computed as the average of the cross-entropy loss over a mini-batch $\mathcal{B}$:
\begin{align}
\mathcal{L} = \frac{1}{|\mathcal{B}|} \sum_{(\bm{x}, \bm{y}) \in \mathcal{B}} \ell_{\mathrm{CE}}(f(\bm{x}), \bm{y})
\end{align}

\subsection{Flooding Regularization}
Flooding regularization~\cite{flood} is a technique designed to suppress overfitting in deep neural networks. In standard training, minimizing the empirical loss toward zero often leads to overfitting to the training data, thereby degrading generalization to unseen test samples. To mitigate this, Ishida et al.~\cite{flood} proposed a method that prevents the mini-batch loss from decreasing below a predefined threshold, referred to as the flooding level $b$. Specifically, the following modified loss function is used:
\begin{align}
\mathcal{L}_{\text{flood}} = \left| \mathcal{L} - b \right| + b
\label{eq:flood}
\end{align}
In this formulation, if the original loss $\mathcal{L}$ is greater than $b$, the optimization proceeds in the standard gradient direction. However, when $\mathcal{L}$ drops below $b$, the gradient reverses its sign, causing the loss to increase. As a result, the loss is maintained around the threshold $b$, which helps suppress overfitting and encourages better generalization.

\subsection{iFlooding Regularization}
iFlooding regularization~\cite{iflood} is an extension of Flooding regularization that applies the flooding effect at the individual sample level rather than over the entire mini-batch. In the original Flooding method, the sign of the gradient is determined based on the relationship between the average mini-batch loss and the flooding level $b$. As a result, even if some samples in the mini-batch are not yet sufficiently learned, the average loss may fall below $b$, causing all gradients to be reversed. This can lead to underfitting, particularly for under-trained samples.
To address this issue, iFlooding regularization applies the flooding operation independently to each sample, ensuring that each is trained appropriately. The loss function is defined as follows:
\begin{align}
\mathcal{L}_{\text{iflood}} = \frac{1}{|\mathcal{B}|} \sum_{(\bm{x}, \bm{y}) \in \mathcal{B}} \left[\left| \ell_{\mathrm{CE}}\left(f(\bm{x}), \bm{y} \right) - b \right| + b \right]
\label{eq:iflood}
\end{align}
The key difference from standard Flooding regularization lies in the order of averaging and regularization: while Flooding applies the operation to the average loss, iFlooding applies it before averaging, at the level of individual samples.

\section{Class-wise Flooding Regularization}
In both standard Flooding and iFlooding regularization, a class-agnostic flooding level $b$ is applied either to the entire mini-batch or to each individual sample. However, in imbalanced datasets, the loss distributions and learning difficulty can vary significantly between majority and minority classes. Using a fixed $b$ across all classes may result in insufficient learning for minority classes, thereby degrading the overall generalization performance.
To address this issue, we propose Class-wise Flooding regularization, which assigns a distinct flooding level $b_k$ to each class, allowing the regularization strength to adapt to the class-specific loss distribution. The Class-wise Flooding loss, denoted by $\mathcal{L}{\mathrm{cwflood}}$, is defined as follows:
\begin{align}
\mathcal{L}{\mathrm{cwflood}} = \frac{1}{|\mathcal{B}|} \sum_{(\bm{x}, \bm{y}) \in \mathcal{B}} \left[ \left| \ell_{\mathrm{CE}}\left(f(\bm{x}), \bm{y}\right) - b(\bm{y}) \right| + b(\bm{y}) \right]
\end{align}
where $\bm{y}$ is a one-hot vector corresponding to class $k$. The class-specific flooding level $b(\bm{y})$ is computed as:
\begin{align}
b(\bm{y}) = \sum_{k=1}^K y_k b_k
\end{align}
where $b_k$ denotes the flooding level assigned to class $k$.
Each $b_k$ is determined based on the relative frequency of samples in class $k$. Let $b_{\mathrm{base}}$ be the base flooding level, $N$ the total number of training samples, and $N_k$ the number of samples in class $k$. Then, the class-wise flooding level is set as:
\begin{align}
b_k = b_{\mathrm{base}} \cdot \frac{N_k}{N}
\end{align}

This setting assigns relatively higher $b_k$ to majority classes, thereby applying stronger regularization to prevent overfitting due to rapid convergence. Conversely, lower $b_k$ values are assigned to minority classes, which helps promote sufficient learning for those classes.

\figtraintestcurvelow

\section{Experiments}
\label{sec:experiment}

\subsection{Experimental Setup}

We evaluated the effectiveness of the proposed method on image classification tasks under imbalanced data with a long-tailed distribution. Following~\cite{ldamloss}, we used \texttt{CIFAR10-LT} and \texttt{CIFAR100-LT} as training datasets. These datasets exhibit long-tailed distributions, where the number of samples per class decays exponentially. The number of samples in class $k$, denoted as $N_k$, is given by:
\begin{align}
N_k = N_{\max} \cdot \left( \frac{1}{\rho} \right)^{\frac{k}{K-1}}.
\end{align}
Here, $N_{\max}$ is the number of samples in the most frequent class, and $\rho$ is the imbalance ratio that controls the degree of imbalance. The number of samples in the rarest class is given by $N_{\min} = N_{\max}/\rho$. We trained models using datasets with imbalance ratios $\rho = \{10, 100\}$, and evaluated classification accuracy on a balanced test set ($\rho = 1.0$). For validation, we used 10\% of the training data.

In all experiments, we used ResNet18~\cite{resnet} as the backbone model. Model parameters were optimized using momentum SGD with a learning rate of 0.01 and a momentum coefficient of 0.9. The batch size is set to 512, and training is conducted for 1,000 epochs. We report classification performance using the best validation model, and provide the mean and standard deviation of test accuracy across three random seeds for statistical robustness. For data augmentation, we applied only \texttt{RandomCrop} and \texttt{RandomHorizontalFlip}.

As baselines, we compared our method with cross-entropy (CE) loss, class-weighted CE loss, focal loss~\cite{focalloss}, Flooding regularization, and iFlooding regularization. For Flooding, iFlooding, and the proposed Class-wise Flooding regularization, we tuned the flooding level from $b = \{0.01, 0.05, 0.1\}$.

\figclasswisetraintestcurvelow
\figclasswisetraintestcurvehigh

\lowimbalancecifarten
\highimbalancecifarten

\subsubsection{Results on \texttt{CIFAR10-LT}}

The classification accuracies on \texttt{CIFAR10-LT} with imbalance ratios $\rho = 10$ and $\rho = 100$ are shown in Tables~\ref{tab:cifar10rho10} and \ref{tab:cifar10rho100}, respectively. Each table presents per-class classification accuracy and the average accuracy across all classes for both the baselines and the proposed method. Bold values indicate the best performance for each class or the overall average.

We observed that the standard cross-entropy loss was significantly affected by the class imbalance, resulting in degraded performance for minority classes. The class-weighted cross-entropy loss improved recognition performance, especially for minority classes. Focal loss~\cite{focalloss}, which is commonly used for imbalanced classification, performed comparably to the standard CE loss. We attribute this to potential overfitting caused by the long training schedule of 1,000 epochs.

Flooding and iFlooding regularization, which have proven effective under class-balanced settings, did not significantly outperform cross-entropy under class-imbalanced conditions. As shown in Table~\ref{tab:cifar10rho100}, their performance on minority classes such as class 8 and class 9 remained low, indicating limited effectiveness in addressing class imbalance.

In contrast, the proposed Class-wise Flooding method achieved consistent improvements in average classification accuracy under both imbalance ratios ($\rho = 10$ and $\rho = 100$), as shown in Tables~\ref{tab:cifar10rho10} and \ref{tab:cifar10rho100}. In particular, the model achieved better recognition of minority classes in the $\rho = 100$ setting. These results demonstrate the effectiveness of assigning class-specific flooding levels to control the learning dynamics for each class.

\imbalancecifarhun

\subsubsection{Results on \texttt{CIFAR100-LT}}

The classification accuracies on \texttt{CIFAR100-LT} with imbalance ratios $\rho = 10$ and $\rho = 100$ are shown in Table~\ref{tab:cifar100}. The table reports the average accuracy across all classes.

Consistent with the results on \texttt{CIFAR10-LT}, we observed that as the imbalance ratio increased from $\rho = 10$ to $\rho = 100$, the proposed method provided generalization performance that is comparable to or better than prior methods. In particular, under $\rho = 100$, our method achieved an improvement of approximately 0.89 points in average accuracy compared to the standard cross-entropy loss.

However, we also found that when both the number of classes and the imbalance ratio increased, the improvement achievable solely through loss-based regularization became limited. This suggests that further investigation is needed on more complex and realistic datasets to fully validate the proposed method under practical conditions.

\subsection{Trends in Average Loss under Flooding}

We analyzed the learning behavior of the proposed method by visualizing the training curves under different flooding levels. Figure~\ref{fig:traintestcurve10} shows the training and validation losses over epochs, averaged over classes, for Flooding, iFlooding, and Class-wise Flooding regularization. As shown in the figure, training loss remained above the specified flooding level in all cases, confirming that the intended effect of Flooding was successfully achieved.

In addition, the validation loss curves demonstrate the regularization effect of Flooding when compared to the cross-entropy baseline (plotted in black). This indicates that all Flooding-based methods were able to suppress overfitting to some extent. However, since the experiments were conducted under class-imbalanced settings, we also observed cases where validation loss increased, particularly for certain hyperparameter choices in standard Flooding regularization.

\subsection{Class-wise Training Dynamics under Flooding}

We further investigated the class-wise training dynamics of the proposed method using per-class loss curves. Figures~\ref{fig:classwisetraintestcurve10} and \ref{fig:classwisetraintestcurve100} show the training and validation losses over epochs for Flooding, iFlooding, and Class-wise Flooding regularization under imbalance ratios $\rho = 10$ and $\rho = 100$, respectively. In these figures, the training loss reflects the regularized loss values, while the validation loss corresponds to the raw cross-entropy loss without any regularization.

From both imbalance settings, we observed a consistent trend: majority classes exhibited faster convergence, while minority classes converged more slowly. This phenomenon became more pronounced as the imbalance ratio increased, which can be attributed to the sampling bias in mini-batch.

For $\rho = 10$, we observed that all regularization methods successfully suppressed overfitting for majority classes, as indicated by their low validation loss. However, overfitting was still evident for minority classes. Among the methods, the proposed Class-wise Flooding showed the smallest increase in loss for both majority and minority classes, suggesting a more balanced regularization effect compared to the other approaches.

On the other hand, under the more severe imbalance ratio of $\rho = 100$, the effectiveness of the proposed method diminished. This is because the class-wise flooding level $b_k$, determined based on sample frequency, became close to zero for minority classes. As a result, while the loss could decrease substantially, the regularization effect of Flooding was not activated. This limitation is also evident from the loss scale in Figure~\ref{fig:classwisetraintestcurve100}.

\bbase

\subsection{Effect of the Base Flooding Level $b_{\mathrm{base}}$}

Finally, we analyzed the effect of the base flooding level $b_{\mathrm{base}}$. Table~\ref{tab:bbase} compares classification accuracy on \texttt{CIFAR10-LT} for different values of $b_{\mathrm{base}}$. The results show that, compared to iFlooding which uses a fixed flooding level across all classes, the proposed class-wise formulation remains effective across both imbalance ratios, regardless of the absolute scale of $b_{\mathrm{base}}$.

While the previous section suggested that increasing the flooding level for minority classes might improve performance, the results in the table indicate that $b_{\mathrm{base}} = 0.1$, which corresponds to higher values for minority classes, does not yield the best performance. This suggests that the effectiveness of class-wise Flooding is not solely determined by the absolute magnitude of the flooding level for minority classes. Instead, these findings support the hypothesis that dynamically adjusting flooding levels based on training dynamics or data distribution could be a promising direction for future work.

\section{Conclusion}

\label{sec:conclusion}

In this study, we proposed Class-wise Flooding, an extension of Flooding regularization that applies class-specific flooding levels, with the goal of mitigating overfitting and improving minority class recognition in multi-class classification tasks with imbalanced data. Experimental results demonstrated that assigning higher flooding levels to fast-converging majority classes effectively suppresses overfitting, while allocating lower flooding levels to slow-converging minority classes allows sufficient training. This strategy proved effective for learning under class-imbalanced settings. However, we also observed that under more extreme imbalance ratios or when the number of classes increases, fixed class-wise flooding levels may become insufficient to properly regulate training. These findings highlight the need for future extensions that adaptively adjust flooding levels according to training dynamics or data distribution. Moreover, applying this approach to larger-scale and more imbalanced datasets remains an important direction for further study.

\section*{Acknowledgements}
This work was supported in part by JSPS KAKENHI Grant Number 25K21226.

\bibliographystyle{splncs04}
\bibliography{main}

\end{document}